\documentclass[letterpaper]{article} 
\usepackage[]{aaai2027}  
\usepackage[hyphens]{url}  
\usepackage{graphicx} 
\usepackage{natbib}  
\usepackage{caption} 
\usepackage{algorithm}
\usepackage{algorithmic}
\usepackage{newfloat}
\usepackage{listings}
\usepackage{amsmath, amssymb}       
\usepackage{booktabs}               
\usepackage{tabularx}    
\usepackage{multirow}   
\DeclareCaptionStyle{ruled}{labelfont=normalfont,labelsep=colon,strut=off} 
\floatstyle{ruled}
\newfloat{listing}{tb}{lst}{}
\floatname{listing}{Listing}
\usepackage{booktabs}

\title{ResonAct: Streaming Metrics for Runtime Diagnosis and Self-Healing in Multi-Agent Systems}

\author{
    Tarun Chintada\textsuperscript{\rm 1,2}\thanks{Work done during an internship at IBM.},
    Neelamadhav Gantyat\textsuperscript{\rm 1},
    Ishaan Romil\textsuperscript{\rm 1,3}\footnotemark[1],
    Renuka Sindhgatta\textsuperscript{\rm 1},
    Soujanya Soni\textsuperscript{\rm 1},
    Sameep Mehta\textsuperscript{\rm 1}
}
\affiliations{
    \textsuperscript{\rm 1}IBM, India\\
    \textsuperscript{\rm 2}Indian Institute of Technology Patna, India\\
     \textsuperscript{\rm 3}International Institute of Information Technology Hyderabad, India\\
    tarunchintada1@gmail.com, neelamadhav@in.ibm.com, ishaan.romil@research.iiit.ac.in, \{renuka.sr, soujanya.soni\}@ibm.com, sameepmehta@in.ibm.com
}

\begin{document}

\maketitle

\begin{abstract}
    Multi-agent systems (MAS) are increasingly used to automate enterprise workflows involving multiple specialized agents, external tools, and long-running task execution. Failures may arise from tool degradation, context propagation errors, coordination breakdowns, or repeated agent interactions that prevent task completion. While existing observability frameworks provide traces and logs, diagnosis and remediation are largely performed after execution completes, limiting opportunities for recovery during runtime. We present ResonAct, a runtime self-healing framework that enables continuous monitoring, diagnosis, and remediation of multi-agent systems through streaming operational metrics. ResonAct ingests execution traces, agent interactions, and tool invocations into a streaming analytics layer that continuously derives task progress, context health, and tool reliability metrics. These metrics serve as runtime control signals for detecting anomalous execution patterns and localizing root causes using a structured failure model. Based on the diagnosed failure, ResonAct dynamically selects remediation policies and performs actions. The framework operates as an external control plane, enabling intervention without modifying application agents or orchestration logic.  We evaluate ResonAct across enterprise workflow scenarios and AppWorld benchmarks. The results show that the streaming metric-based analysis identifies execution degradations and localizes faults. Furthermore, policy-driven remediation improves task completion rates by up to 10.00 percentage points, with detection precision ranging from 70.59\% to 82.91\%, recall from 63.09\% to 100\%, recovery rates from 10.48\% to 46.67\%, and runtime overhead ranging from $-0.25\%$ to 14.12\% across the evaluated configurations.
\end{abstract}

\section{Introduction}

Large Language Model (LLM)-based multi-agent systems (MAS) are increasingly being used to automate complex workflows, such as software development, tool orchestration, and data-intensive tasks~\cite{qian2024communicativeagentssoftware, luo2026dataagentslevelsstate}. By coordinating specialized agents with external tools over extended execution horizons, these systems can perform tasks that require multiple interdependent decisions and actions. However, increasing autonomy also introduces reliability challenges. Failures in MAS can arise not only from conventional software faults, but also from interactions among agents, tool failures, context propagation errors, coordination breakdowns, and deviations from the intended execution trajectory~\cite{cemri2025why, huang2025resiliencemas}.

Recent empirical work provides evidence that failures in MAS are both diverse and systematic~\cite{cemri2025why}.
Failures arise due to execution-time behaviors such as step repetition, loss of conversation history, reasoning-action mismatch, task derailment, and premature termination. 
Existing operational approaches provide mechanisms for observing and recording such executions through traces, logs, and other telemetry~\cite{moshkovich2025taming}. However, failure detection and root cause analysis are often carried out after an execution has failed or produced an undesirable outcome~\cite{solomon2025lumimas}. This post-hoc paradigm is particularly problematic for long-running enterprise agentic workflows, where anomalous behavior can incur substantial computational cost. Recent work has also shown that agentic workflows can incur substantial and highly variable token consumption as a result of recursive behavior, context growth, and repeated tool interactions, with large cost differences possible even for identical inputs~\cite{bai2026aiagentsspendmoney}. Consequently, post-execution analysis can result in unnecessary computational expenditure and increased execution latency. Hence, conventional observability mechanisms primarily provide visibility into system behavior but do not provide a general mechanism for intervening in an ongoing agent execution.

To address these challenges, we present \textbf{ResonAct} (Resonate and Act), a runtime self-healing framework for LLM-based multi-agent systems. ResonAct continuously ingests execution traces, agent interactions, and tool invocations through a streaming analytics layer and derives dynamic health signals, including task progress, context reliability, and tool health. These signals are used to identify emerging execution anomalies and localize potential failure causes. Policy-driven remediation actions are then applied during execution, enabling automated intervention without requiring modifications to individual agent implementations. ResonAct further separates monitoring, failure analysis, and remediation policies from the application agents and orchestrators through an external control plane.

\noindent\textbf{Contributions:} To summarize, we make the following contributions:

(1) We propose \textit{ResonAct}, a runtime self-healing framework for LLM-based multi-agent systems that enables continuous failure detection, diagnosis, and policy-driven remediation during task execution.

(2) We introduce a \textit{streaming metrics-based control mechanism} that transforms execution telemetry, including agent interactions, tool invocations, and execution traces, into metrics representing task progress, context health, tool reliability, and execution state. These signals enable detection and localization of emerging failures before task termination.

(3) We develop a \textit{policy-driven remediation mechanism} that maps diagnosed failure conditions to targeted runtime interventions, enabling the system to recover from execution failures while preserving the underlying task objective.

(4) We evaluate ResonAct across enterprise workflow scenarios and the AppWorld benchmark, measuring failure detection, task completion and recovery, and runtime overhead. Our evaluation shows improved task completion and recovery while maintaining low operational overhead.

\section{Related Work}

We position ResonAct within prior work on multi-agent systems, MAS failure analysis, and runtime observability and self-healing.

\subsection{Multi-Agent Systems}

LLM-based multi-agent frameworks provide abstractions for agent specialization, communication, task decomposition, and workflow orchestration. Frameworks including AutoGen~\cite{wu2024autogen}, CAMEL~\cite{li2023camel}, MetaGPT~\cite{hong2024metagpt}, CrewAI~\cite{crewai2024}, and LangGraph~\cite{wang2024agentailanggraphmodular} support increasingly complex agentic workflows. While these frameworks provide mechanisms for coordinating agents and executing tasks, runtime reliability is generally not their primary focus. In particular, they do not provide general mechanisms for continuously detecting failures, diagnosing their causes, and applying automated recovery actions during execution.

\subsection{Failure Analysis and Reliability}

Recent work has established that MAS failures are diverse and arise from interactions among agents, system components, and execution state. Multi-Agent System Failure Taxonomy (MAST) provides an empirically grounded taxonomy of 14 failure modes, developed from 150 execution traces and evaluated on 1,600 traces across seven frameworks~\cite{cemri2025why}. 
Complementary work studies failure propagation, fault injection, memory inconsistency, and agent collaboration reliability~\cite{lin-etal-2026-agentask, jia2026masfirefaultinjectionreliability, wang2023augmentinglanguagemodelslongterm, zhang2024llmcascade}. Collectively, these studies demonstrate that MAS reliability requires systematic failure detection and diagnosis rather than isolated application-level fixes. However, they primarily focus on characterizing, analysing, or injecting failures rather than providing mechanisms for automated runtime recovery.

\subsection{Observability and Self-Healing Systems}

Observability provides visibility into system behavior through logs, metrics, traces, and events~\cite{itobservability}. AIOps extends observability with automated fault detection and root cause analysis, demonstrating the value of closed-loop management for deterministic software and infrastructure~\cite{pei2025flowofaction}. However, LLM-based MAS introduces additional failure modes involving agent reasoning, coordination, context propagation, and tool-mediated execution that are not directly addressed by traditional infrastructure-oriented approaches.

Recent work addresses runtime observability specifically for MAS. AgentOps introduces a pipeline spanning observation, metric collection, issue detection, root-cause analysis, and optimization, highlighting the need for runtime intervention in agentic workflows~\cite{moshkovich2025taming, alsayyad2026agenttraces}. LumiMAS provides platform-agnostic runtime monitoring using an LSTM autoencoder for anomaly detection, followed by failure classification and root-cause analysis using the MAST taxonomy~\cite{solomon2025lumimas}. SentinelAgent models MAS execution as a dynamic interaction graph and combines graph-based anomaly detection with an LLM-powered oversight agent for runtime analysis and intervention~\cite{he2025sentinelagent}.

Existing work demonstrates the feasibility of runtime monitoring for MAS, but differs from ResonAct in its detection and intervention mechanisms. 
\textit{ResonAct} uses \textit{interpretable streaming operational metrics} as runtime control signals and does not require model training or per-step LLM judgement. 

\begin{figure*}[tb]
    \centering
    \includegraphics[
        width=0.95\textwidth,
        height=7.5cm
    ]{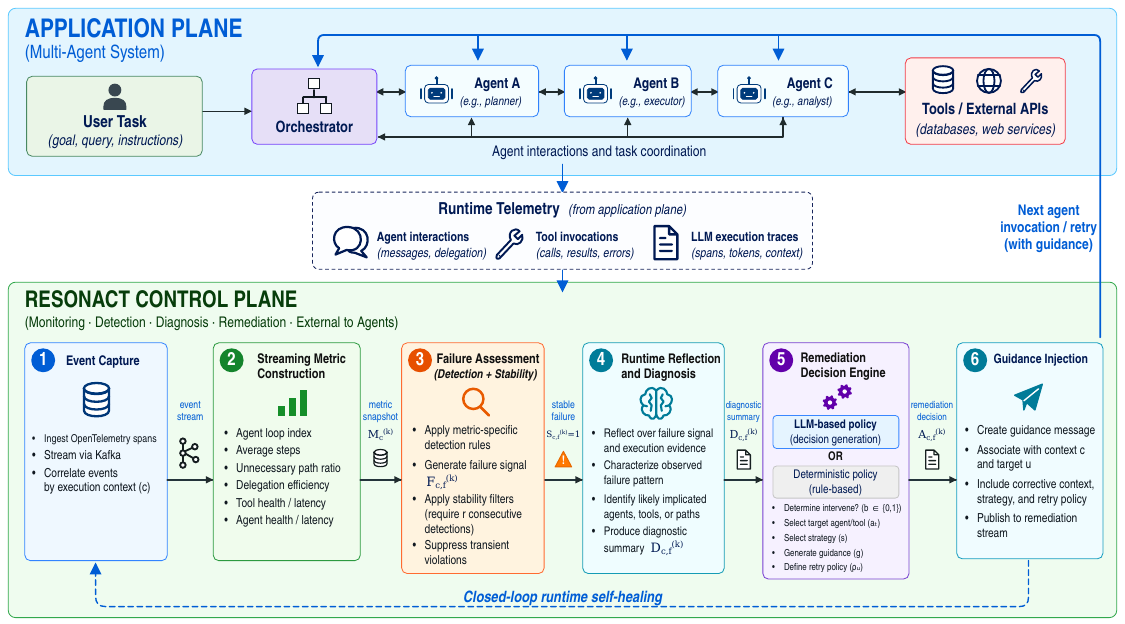}
    \caption{Architecture of the proposed ResonAct framework.}
    \label{fig:architecture}
\end{figure*}

\section{The ResonAct Framework} \label{sec:framework}

ResonAct operates as an external control plane alongside application agents and their orchestrator, separating runtime monitoring, failure analysis, and remediation policies from task-specific agent reasoning and tool implementations. As illustrated in Figure~\ref{fig:architecture}, ResonAct forms a closed control loop comprising: (1) runtime event capture, (2) streaming metric construction, (3) failure detection, (4) stability filtering, (5) remediation selection, and (6) corrective guidance injection. The components communicate through an event stream and can therefore operate independently of the underlying agent framework.

\subsection{Failure Taxonomy and Runtime Events Capture}

ResonAct builds on existing empirical characterizations of MAS failures~\cite{cemri2025why} and organizes runtime conditions into three operational categories:
\begin{itemize}
\item\textbf{Common Failures:} transient errors, timeouts, and retry failures;
\item\textbf{Silent Failures:} context decay, state divergence, redundant execution, and other degradations without explicit errors;
\item\textbf{Critical Failures:} coordination breakdowns, unbounded loops, and cascading failures threatening task completion or resource bounds;
\end{itemize}

Additionally, three principal forms of execution evidence: agent interactions, tool invocations, and LLM execution traces are used. In the current implementation, events are generated through an observability platform\footnote{\url{https://langfuse.com/}} and forwarded to an event-streaming platform\footnote{\url{https://kafka.apache.org/}}. 

\subsection{Streaming Metric Construction} \label{sec:context-builder}
Individual runtime events are often insufficient to determine whether an agent workflow is progressing normally or approaching a failure state. Hence, the event stream is processed to incrementally compute metrics that summarize execution behavior. An execution context $c$ groups the runtime events associated with one attempt to fulfil a user request, including executions that complete, fail, time out, or are otherwise terminated. It defines the scope over which runtime metrics are computed. The metric state for context $c$ at time $t$ is \begin{equation} 
\mathbf{M}_c(t) = \left[m_1(c,t),m_2(c,t),\ldots,m_K(c,t)\right], \label{eq:metric-state} 
\end{equation} 
where $K$ is the number of available runtime metrics and each $m_j(c,t)$ is computed from the events observed for $c$ up to time $t$. 

The metrics are updated incrementally as events arrive and are organized along two axes. The first specifies the \emph{unit of analysis}: query-level metrics summarize the state of an entire execution context, agent-level metrics characterize an individual agent, and tool-level metrics characterize the reliability and performance of an external tool. The second specifies the \emph{computation mechanism}: count-based metrics aggregate structured runtime events, embedding-based metrics compare semantic representations, and LLM-as-judge metrics assess properties that cannot be derived directly from structured telemetry. Query-, agent-, and tool-level metrics are updated continuously as runtime events arrive and are evaluated at configurable intervals, such as every \(x\) seconds or after a configurable number of execution steps. Table~\ref{tab:metric_failure_map} summarizes the metrics used to construct the runtime failure signatures. Metric computation is independent of remediation policy. A metric describes observed execution behavior but does not determine whether an intervention should occur. This separation allows metrics to be reused across failure detectors and enables remediation policies to change without modifying the streaming analytics layer. Detailed metric definitions are provided in the Appendix.

\begin{table}[tb]
\centering
\resizebox{\columnwidth}{!}{%
\begin{tabular}{l l l l}
\toprule
\textbf{Category} & \textbf{Failure Mode} & \textbf{Driving Metric} & \textbf{Detection Rule} \\
\midrule
\multirow{3}{*}{\textbf{Critical}}
  & agent\_loop
  & agent\_loop\_index
  & $\geq \tau_L$ \\
  & unbounded\_loop
  & average\_steps
  & $> P_{95}$ \\
  & task\_incomplete
  & task\_completion\_rate
  & $< \tau_C$ \\
\midrule
\multirow{3}{*}{\textbf{Silent}}
  & redundant\_execution
  & unnecessary\_path\_ratio
  & $\geq \tau_U$ \\
  & context\_propagation\_gap
  & delegation\_efficiency
  & $< \tau_D$ \\
  & decision\_repetition
  & agent\_loop\_index
  & recurrent \\
\midrule
\multirow{3}{*}{\textbf{Common}}
  & tool\_degradation
  & tool\_health / latency
  & fail-rate $\geq \tau_T$ \\
  & agent\_degradation
  & agent\_health / latency
  & fail-rate $\geq \tau_A$ \\
  & API call drift & call\_efficiency & Moving-Avg. ratio $\geq \tau_P$ \\
\bottomrule
\end{tabular}%
}
\caption{Metric-driven runtime failure signatures considered.}
\label{tab:metric_failure_map}
\end{table}

\subsection{Failure Assessment} \label{sec:failure-assessment} 
At evaluation cycle $k$, occurring at time $t_k$, the Failure Assessment component reads the latest metric snapshot and applies each failure-specific detector $d_f$: 
\begin{equation} 
\mathbf{M}_c^{(k)}=\mathbf{M}_c(t_k), \qquad d_{c,f}^{(k)}=d_f\!\left(\mathbf{M}_c^{(k)}\right)\in\{0,1\}. \label{eq:failure-detection} 
\end{equation} 
The evaluation cadence is configurable and may be defined using a wall-clock interval or a specified number of execution steps.
The detection conditions are defined in Table~\ref{tab:metric_failure_map}. If $d_{c,f}^{(k)}=1$, the component constructs the candidate failure signal
\begin{equation} 
F_{c,f}^{(k)} = \left(f,c,\mathbf{M}_c^{(k)}\right), \label{eq:failure-signal} 
\end{equation} which identifies the failure type and execution context and retains the supporting metric evidence. To suppress transient violations, a failure must be detected for $r$ consecutive evaluation cycles: 
\begin{equation} 
S_{c,f}^{(k)} = \prod_{\ell=0}^{r-1}d_{c,f}^{(k-\ell)}. \label{eq:stability} 
\end{equation} 
Thus, $S_{c,f}^{(k)}=1$ only when the failure is detected in the current cycle and the preceding $r-1$ cycles. The corresponding signal $F_{c,f}^{(k)}$ is then treated as stable and forwarded for further diagnosis.  The persistence parameter $r$ controls the trade-off between early intervention and resistance to transient detections.

\subsection{Runtime Reflection and Diagnosis} \label{sec:diagnosis} 
When $S_{c,f}^{(k)}=1$, the stable failure signal $F_{c,f}^{(k)}$ is passed to the diagnosis component. Because an expected execution path is not generally available at runtime, an LLM reflects over the failure signal and available execution evidence, including recent agent interactions, tool outcomes, execution history, and the current metric state. It produces 
\begin{equation} 
D_{c,f}^{(k)}=(q,e), \label{eq:diagnostic-output} 
\end{equation} where $q$ characterizes the observed failure pattern and $e$ summarizes the supporting evidence, including likely implicated agents, tools, or execution paths. This output represents an evidence-based interpretation, not a definitive root-cause determination.

\subsection{Remediation Decision Engine} \label{sec:decision-engine} 
The Remediation Decision Engine consumes the stable failure signal $F_{c,f}^{(k)}$ and diagnostic output $D_{c,f}^{(k)}$ to produce \begin{equation} 
A_{c,f}^{(k)} = \left(b,u,s,g,\rho_a\right), \label{eq:remediation-action} 
\end{equation} where $b\in\{0,1\}$ indicates whether to intervene, $u$ identifies the target agent or tool, $s$ specifies the remediation strategy, $g$ contains the corrective guidance, and $\rho_a\in\{0,1\}$ indicates whether a retry is authorized. The engine supports LLM-based and deterministic remediation policies. The LLM-based policy generates a context-specific decision from the diagnostic output, while the deterministic policy maps recognized failure types to predefined, bounded actions. Deterministic remediation is also used when the LLM is unavailable or fails to generate the output. Guidance is published for runtime injection only when $b=1$.

\subsection{Runtime Guidance Injection} \label{sec:guidance-injection} 
The remediation decision $A_{c,f}^{(k)}$ is published as a guidance message on a dedicated remediation stream (when $b=1$). An agent-side integration component retrieves guidance associated with execution context $c$ and incorporates it into the applicable agent's prompt before the next invocation or retry. The guidance may instruct the agent to avoid a failing execution path, invoke an alternative agent or tool, recover missing context, or restrict further retries. ResonAct does not modify an LLM generation already in progress; intervention occurs only at an \emph{agent invocation boundary}.

\section{Experiments}
We evaluate ResonAct framework by answering the following three research questions:

\noindent\textbf{RQ1: Detection.}
Can streaming metrics reliably detect multi-agent failures during task execution?

\noindent\textbf{RQ2: Remediation.}
Does policy-driven remediation improve task completion and recovery from detected failures?

\noindent\textbf{RQ3: Efficiency.}
Can task reliability improve while maintaining low runtime and resource overhead?

\noindent\textbf{Models:} We use two compact, open-weight LLMs, \texttt{granite-4.1-8b} (Granite-8B)\footnote{\url{https://huggingface.co/ibm-granite/granite-4.1-8b}} and \texttt{qwen3-8b} (Qwen3-8B)\footnote{\url{https://huggingface.co/Qwen/Qwen3-8B}}, to evaluate the framework. We focus intentionally on 8B-scale models because they are more representative of practical industry deployments, where constraints on latency, inference cost, memory footprint, and on-premises serving often make larger models less feasible. This enables us to assess the effectiveness of the proposed architecture and remediation framework under realistic resource-constrained deployment settings.

\subsection{Evaluation Metrics} \label{sec:evaluation-metrics}

Table~\ref{tab:metrics} defines the metrics used for RQ1--RQ3. Detection is evaluated against independently labelled execution outcomes. A task is considered recovered only when a detected failed execution completes successfully after remediation. 

\begin{table}[t] 
\centering 

\scriptsize \setlength{\tabcolsep}{4pt}
\renewcommand{\arraystretch}{0.95} 
\begin{tabular}{@{}llp{0.73\columnwidth}@{}} 
\hline \textbf{RQ} & \textbf{Metric} & \textbf{Definition} \\ \hline 
RQ1 & Recall & Labelled failed executions detected / all labelled failed executions \\
            & Precision & Correctly detected failed executions / all detected executions \\ 
            & FPR & Labelled healthy executions detected as failures / all labelled healthy executions \\ \hline
RQ2 & Completion & Successfully completed tasks / all evaluated tasks \\ 
    & Recovery & Failed executions completed after remediation / detected failed executions \\ \hline 
RQ3 & Runtime & Relative latency increase over the corresponding baseline \\ 
    & CPU & Change in CPU utilization relative to the corresponding baseline \\ 
    & Memory & Change in memory utilization relative to the corresponding baseline \\ \hline 
    \end{tabular} 
\caption{Evaluation metrics for detection, recovery, overhead.} 
\label{tab:metrics} 
\end{table}

\subsection{Experimental Setup}
We present our results on two multi-agent system settings.
\paragraph{AppWorld MAS} The original AppWorld benchmark employs a single ReAct agent that independently reasons over tasks and interacts directly with all application APIs within a unified conversation loop. We extend this architecture to a distributed multi-agent system (MAS) comprising an OrchestratorAgent and nine application-specific worker agents. The orchestrator delegates app-related subtasks to specialized agents, each restricted to a single application, provisioned with the required access credentials, and equipped with application-specific API knowledge. Agent-to-environment interactions are handled through a traced execution layer that enables efficient tool attribution. 

All experiments were performed using the benchmark's test\_normal evaluation split, providing a standardized and consistent basis for comparing the performance of the MAS and MAS+ResonAct configurations.

\paragraph{Sales MAS}
The setup consists of ten agents that model an end-to-end B2B sales fulfillment workflow. A Host Agent orchestrates execution by decomposing natural-language requests and routing subtasks to specialized agents (via the A2A protocol), passing intermediate outputs between agents until all requested business functions are completed. The system comprises ten domain-specific agents covering order management, inventory tracking, payments, order fulfillment, warehouse operations, pricing, vendor management, procurement, shipping, and returns, each encapsulating its own tools and business logic and exposed as an independent HTTP service. 

The validation dataset is a 600-query ground-truth benchmark spanning all ten domain agents, with queries evenly split across Medium, Difficult, and Complex difficulty tiers. It was deliberately engineered so that approximately 60\% of the queries represent failure-oriented stress scenarios by incorporating policy edge cases, missing context, and multi-agent dependency chains, making it the canonical evaluation set for measuring system correctness under realistic stress conditions.

\subsubsection{MAS with ResonAct}
To improve robustness, we integrate the ResonAct remediation layer into the AppWorld and Sales use case, where streamed agent events are analyzed to detect persistent failures and automatically generate corrective guidance that is injected into both worker and orchestrator reasoning loops, enabling adaptive self-healing while remaining fully optional and preserving the baseline MAS when disabled.

\subsection{Results and Discussion}
\label{sec:results}

We evaluate ResonAct along three dimensions: (RQ1) the reliability of its
metric-driven failure detector, (RQ2) its effect on end-to-end task completion
and recovery, and (RQ3) the runtime cost introduced by continuous diagnosis
and remediation. We evaluate detection, remediation, and efficiency end-to-end across
Sales MAS and AppWorld.


\begin{figure}[t]
    \centering
    \includegraphics[
        width=0.95\columnwidth,
        height=3.7cm
    ]{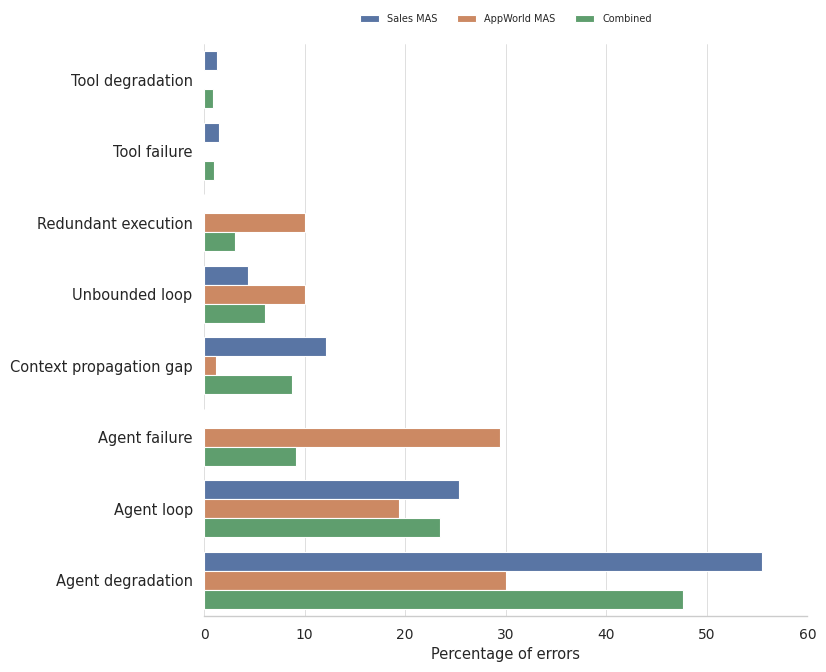}
    \caption{Operational failure-type distribution across Sales MAS and AppWorld MAS.}
    \label{fig:operational-failure-classes}
\end{figure}

\subsubsection{RQ1: Runtime Failure Detection}

Detection performance varies across workloads and model backends.
On Sales MAS, Granite-8B achieves 76.21\% precision and 69.15\% recall,
while Qwen3-8B achieves 72.11\% precision and 63.09\% recall. The
corresponding F1 scores are 72.51\% and 67.30\%.

AppWorld has a greater coverage of failures. Recall reaches 100\% for both
models, with precision of 82.91\% for Granite-8B and 70.59\% for Qwen3-8B.
However, this sensitivity is accompanied by elevated false-positive rates,
particularly for Granite-8B (72.97\%). These results expose a
sensitivity-selectivity trade-off: ResonAct captures a larger fraction of
failures on AppWorld but can also intervene on healthy executions.

Figure~\ref{fig:operational-failure-classes} further shows that failure
composition differs substantially across workloads. Agent degradation and
agent loops constitute the dominant combined failure classes, whereas
AppWorld contains a comparatively larger proportion of agent failures.
This variation helps explain why a single detector configuration does not
behave identically across benchmarks.

\textbf{RQ1 Summary.}
ResonAct detects failures across both workloads and model families, with
precision ranging from 70.59 - 82.91\% and recall from 63.09 - 100\%.
The variation in FPR indicates that workload-aware detector calibration
remains important.


\begin{table*}[t]
\centering
\small
\setlength{\tabcolsep}{3.5pt}
\renewcommand{\arraystretch}{1.05}

\begin{tabular}{@{}l|l|rrrr|rr|rr|rr|r@{}}
\hline
\textbf{Benchmark} & \textbf{Model} &
\multicolumn{4}{c|}{\textbf{Detection (\%)}} &
\multicolumn{2}{c|}{\textbf{Completion (\%)}} &
\textbf{Gain} & \textbf{Recovery} &
\multicolumn{2}{c|}{\textbf{Avg Time (s)}} &
\textbf{Overhead} \\
\cline{3-6}
\cline{7-8}
\cline{11-12}
& & \textbf{Precision} & \textbf{Recall} & \textbf{F1} & \textbf{FPR} &
\textbf{Base} & \textbf{ResonAct} &
\textbf{(pp)} & \textbf{(\%)} &
\textbf{Base} & \textbf{ResonAct} &
\textbf{(\%)} \\
\hline

\multirow{2}{*}{AppWorld MAS}
 & Granite-8B
 & 82.91 & 100.00 & 90.65 & 72.97
 & 2.01 & \textbf{7.14}
 & +5.13 & 32.06
 & 122.32 & 122.01 & -0.25 \\

 & Qwen3-8B
 & 70.59 & 100.00 & 82.76 & 41.67
 & 11.31 & \textbf{16.07}
 & +4.76 & 10.48
 & 143.54 & 163.82 & +14.12 \\
\hline

\multirow{2}{*}{Sales MAS}
 & Granite-8B
 & 76.21 & 69.15 & 72.51 & 53.83
 & 80.83 & \textbf{88.00}
 & +7.17 & 46.67
 & 22.43 & 23.70 & +5.66 \\

 & Qwen3-8B
 & 72.11 & 63.09 & 67.30 & 51.18
 & 73.67 & \textbf{83.67}
 & +10.00 & 37.97
 & 23.58 & 24.06 & +2.04 \\
\hline

\end{tabular}

\caption{End-to-end detection, task completion, recovery, and runtime
performance. Runtime overhead is computed relative to the corresponding
baseline execution time.}
\label{tab:remediation}
\end{table*}

\subsubsection{RQ2: Runtime Remediation and Task Completion}

As shown in Table~\ref{tab:remediation}, ResonAct improves task completion
across all four benchmark--model configurations. The largest gain occurs on
Sales MAS with Qwen3-8B, increasing from 73.67\% to 83.67\% (+10.00 pp),
while Granite-8B improves from 80.83\% to 88.00\% (+7.17 pp). Recovery among
detected failed executions reaches 37.97\% and 46.67\%, respectively.

AppWorld also shows positive gains. Granite-8B improves from 2.01\% to
7.14\% (+5.13 pp), with 32.06\% recovery, while Qwen3-8B improves from
11.31\% to 16.07\% (+4.76 pp), with 10.48\% recovery. The lower recovery
rates indicate that longer, tool-intensive AppWorld trajectories are harder
to repair than the Sales workflows.

Post-trigger execution is substantially more reliable. In the Granite-8B
Sales run, 43 of 44 initiated top-level retries succeed (97.7\%), including
7/7 Medium, 19/20 Difficult, and 17/17 Complex cases. This distinguishes
retry success from recovery: the former measures executions where a retry is
actually initiated, whereas recovery considers all detected failed executions.

\textbf{RQ2 Summary.}
ResonAct improves completion by 4.76--10.00 percentage points across all
evaluated configurations, with stronger recovery on Sales than AppWorld.

\subsubsection{RQ3: Runtime and Resource Overhead}

ResonAct introduces additional computation through monitoring, diagnosis,
and remediation. On Sales MAS, mean execution time increases from 22.43~s to
23.70~s for Granite-8B (+5.66\%) and from 23.58~s to 24.06~s for Qwen3-8B
(+2.04\%). On AppWorld, Granite-8B remains essentially unchanged
(122.32~s vs.\ 122.01~s, $-0.25\%$), whereas Qwen3-8B increases from
143.54~s to 163.82~s (+14.12\%).

Resource overhead remains modest. Across the Sales runs, aggregate CPU
utilisation increases by approximately 4.84\%, while memory consumption
decreases by approximately 2.55\%. For AppWorld with Qwen3-8B, CPU
utilisation increases by 4.18 percentage points and RAM utilisation by
2.33 percentage points, while mean physical RAM usage decreases by 3.36\%.
Thus, the additional cost is driven primarily by longer execution and
increased compute activity rather than sustained memory growth.

The overhead should be considered together with the reliability gains:
Sales achieves completion improvements of 7.17-10.00 pp with only
2.04–5.66\% additional runtime, while AppWorld shows greater
model-dependent variation.

\begin{figure}[t]
    \centering
    \includegraphics[width=0.9\columnwidth, height=3.4cm]{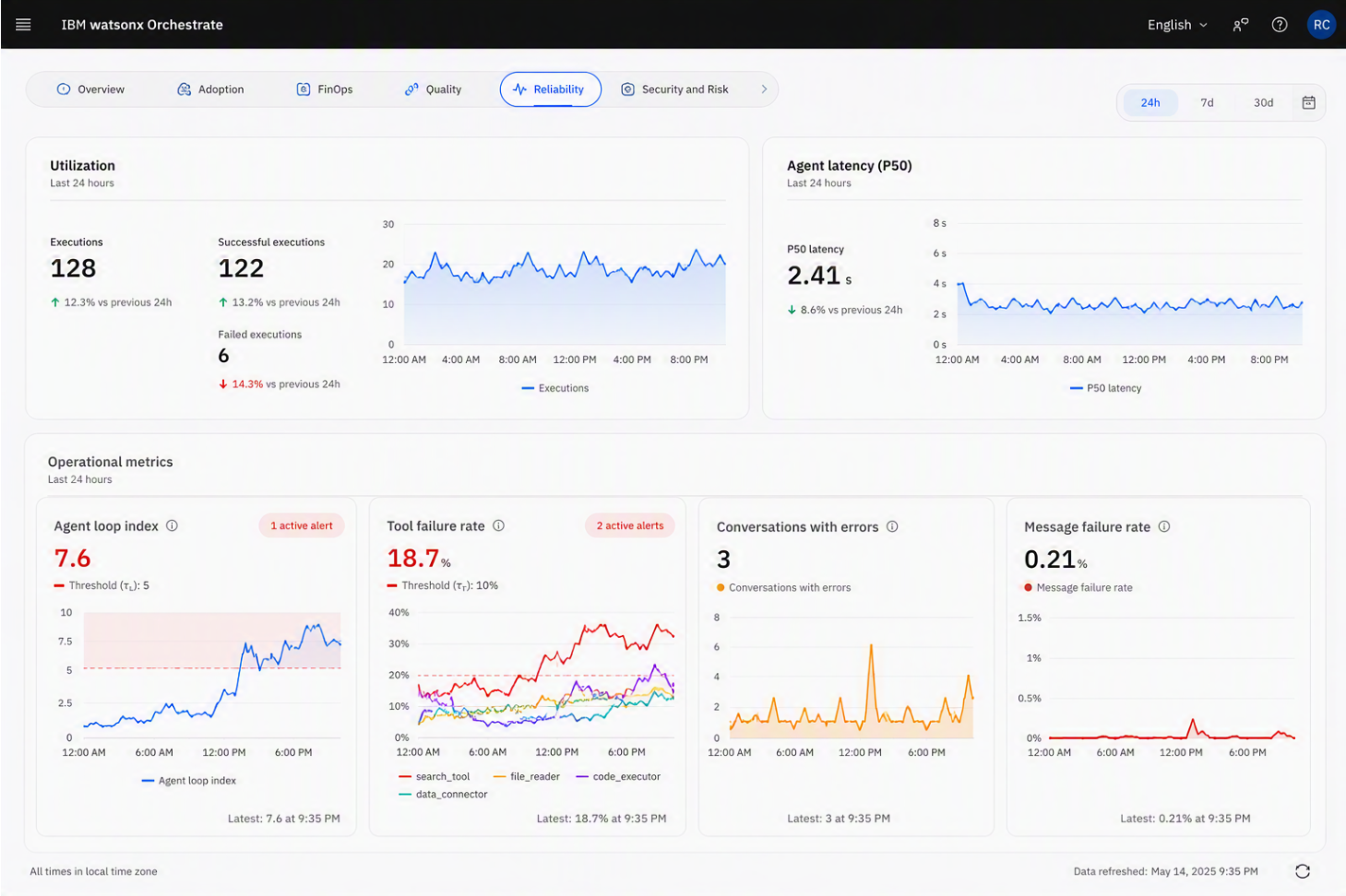}
    \caption{Dashboard reporting failures through ResonAct.}
    \label{fig:wxo-dashboard}
\end{figure}

\textbf{RQ3 Summary.}
Runtime overhead ranges from negligible to 14.12\%, while CPU and memory
changes remain comparatively modest.

\paragraph{Overall Implications.}
ResonAct consistently improves task completion while exposing a clear
trade-off between detection quality, recovery difficulty, and runtime cost.
Recovery is stronger on the structured Sales workload, whereas AppWorld's
longer and more heterogeneous trajectories make both remediation and
overhead more variable. The high post-trigger retry success further suggests
that failure selection and calibration are currently more limiting than the
execution of remediation itself. These findings support workload-aware
detector calibration and selective intervention policies that apply
additional computation when recovery is likely to improve the final outcome.

\begin{figure}[t]
    \centering
    \includegraphics[
        width=0.75\columnwidth,
        height=2.7cm
    ]{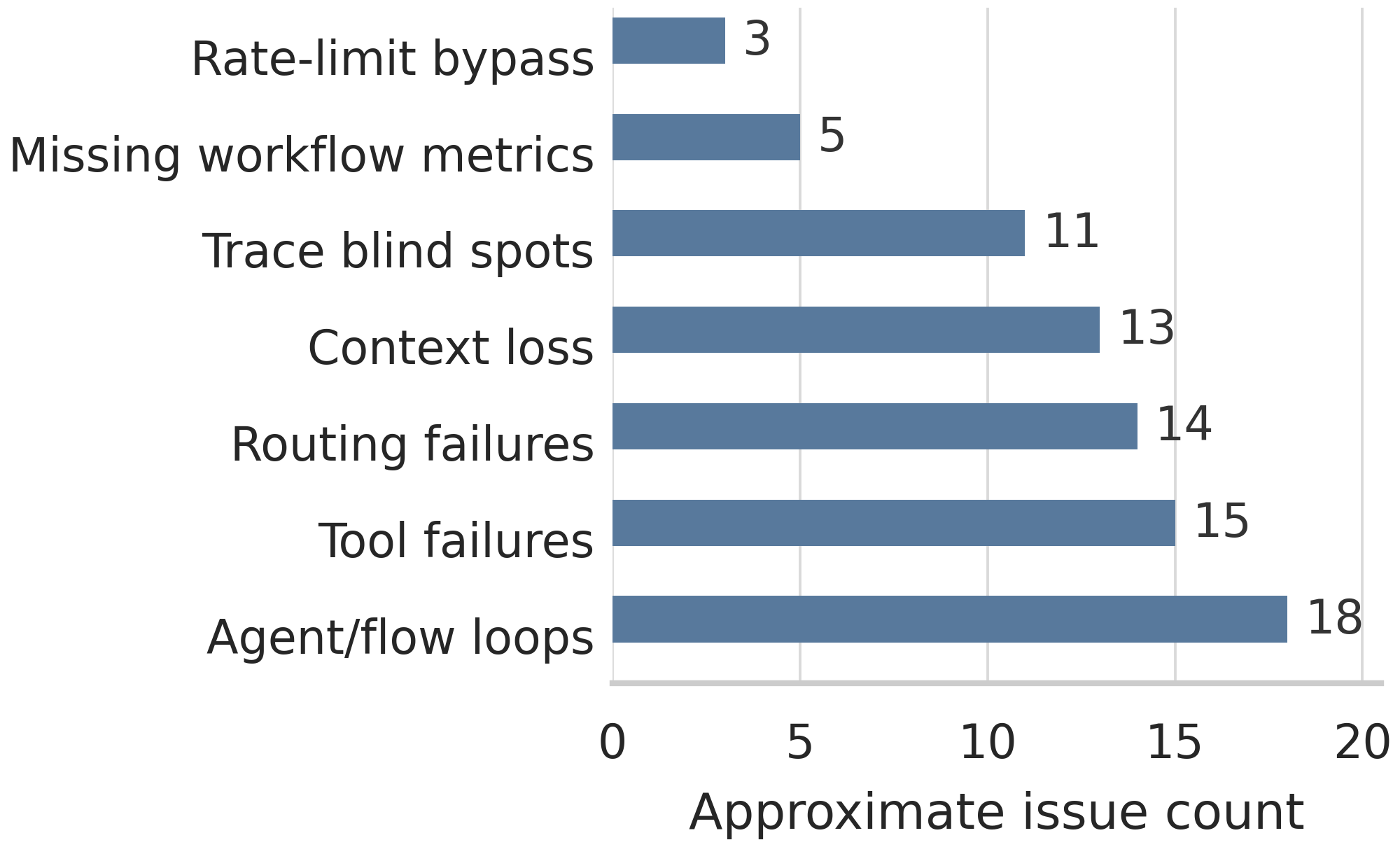}
    \caption{Distribution of identified operational failure classes}
    \label{fig:operational-failure}
\end{figure}

\subsection{Deployment Plan} \label{sec:deployment} 
ResonAct is integrated with IBM watsonx Orchestrate (WXO)\footnote{\url{https://www.ibm.com/products/watsonx-orchestrate}} as a technical preview for streaming runtime diagnosis in an enterprise agent platform. The operational motivation is based on an analysis of 904 issues from an internal WXO support repository, including issue descriptions and discussion threads. Approximately 160 issues were mapped to capabilities involving runtime event capture, metric computation, failure detection, or remediation. Recurring classes included agent and flow loops, intermittent tool failures, non-deterministic routing, context-propagation gaps, and incomplete traces (Figure~\ref{fig:operational-failure}). This analysis establishes the operational relevance of the targeted failure classes. In the current detection-only preview, WXO runtime events are streamed through Confluent Kafka\footnote{\url{https://www.confluent.io/cloud-kafka/}} to ResonAct's metric-construction and failure-detection components. The implemented integration incrementally computes the agent-loop index and per-tool failure rates, enabling potential execution loops and sustained tool degradation to be identified during execution.  Figure~\ref{fig:wxo-dashboard} shows how the resulting signals are surfaced in the technical-preview dashboard.

Progression beyond the current preview is planned with a controlled evaluation of remediation effectiveness and operational overhead. In particular, the latency, token consumption, and cost of LLM-based decision generation will be weighed against improvements in task recovery for pilot customers. Because remediation can alter an execution path and impact the task outcome, its usefulness will also be assessed with platform users. The evidence through pilot customers will guide the safeguards of the deployment.

\section{Conclusion}
We presented ResonAct, an external control plane that derives continuous operational metrics from multi-agent execution events and uses them to drive failure-specific runtime interventions. Its streaming architecture enables detection and remediation at agent invocation boundaries without embedding control logic in individual agents. Integration with the agent runtime platform provides a technically grounded path to controlled operational deployment.

\bibliography{aaai2027}

@inproceedings{huang2025resiliencemas,
  title={On the Resilience of {LLM}-Based Multi-Agent Collaboration with Faulty Agents},
  author={Huang, Jen-Tse and Zhou, Jiaxu and Jin, Tailin and Zhou, Xuhui and Chen, Zixi and Wang, Wenxuan and Yuan, Youliang and Lyu, Michael R. and Sap, Maarten},
  booktitle={Proceedings of the 42nd International Conference on Machine Learning},
  pages={26202--26226},
  year={2025},
  volume={267},
  series={Proceedings of Machine Learning Research},
  publisher={PMLR},
  url={https://proceedings.mlr.press/v267/huang25ay.html}
}

@misc{luo2026dataagentslevelsstate,
      title={Data Agents: Levels, State of the Art, and Open Problems}, 
      author={Yuyu Luo and Guoliang Li and Ju Fan and Nan Tang},
      year={2026},
      eprint={2602.04261},
      archivePrefix={arXiv},
      primaryClass={cs.DB},
      url={https://arxiv.org/abs/2602.04261}, 
}

@misc{qian2024communicativeagentssoftware,
      title={ChatDev: Communicative Agents for Software Development}, 
      author={Chen Qian and Wei Liu and Hongzhang Liu and Nuo Chen and Yufan Dang and Jiahao Li and Cheng Yang and Weize Chen and Yusheng Su and Xin Cong and Juyuan Xu and Dahai Li and Zhiyuan Liu and Maosong Sun},
      year={2024},
      eprint={2307.07924},
      archivePrefix={arXiv},
      primaryClass={cs.SE},
      url={https://arxiv.org/abs/2307.07924}, 
}

@inproceedings{
wu2024autogen,
title={AutoGen: Enabling Next-Gen {LLM} Applications via Multi-Agent Conversations},
author={Qingyun Wu and Gagan Bansal and Jieyu Zhang and Yiran Wu and Beibin Li and Erkang Zhu and Li Jiang and Xiaoyun Zhang and Shaokun Zhang and Jiale Liu and Ahmed Hassan Awadallah and Ryen W White and Doug Burger and Chi Wang},
booktitle={First Conference on Language Modeling},
year={2024},
url={https://openreview.net/forum?id=BAakY1hNKS}
}

@inproceedings{
li2023camel,
title={{CAMEL}: Communicative Agents for ''Mind'' Exploration of Large Language Model Society},
author={Guohao Li and Hasan Abed Al Kader Hammoud and Hani Itani and Dmitrii Khizbullin and Bernard Ghanem},
booktitle={Thirty-seventh Conference on Neural Information Processing Systems},
year={2023},
url={https://openreview.net/forum?id=3IyL2XWDkG}
}

@inproceedings{
hong2024metagpt,
title={Meta{GPT}: Meta Programming for A Multi-Agent Collaborative Framework},
author={Sirui Hong and Mingchen Zhuge and Jonathan Chen and Xiawu Zheng and Yuheng Cheng and Jinlin Wang and Ceyao Zhang and Zili Wang and Steven Ka Shing Yau and Zijuan Lin and Liyang Zhou and Chenyu Ran and Lingfeng Xiao and Chenglin Wu and J{\"u}rgen Schmidhuber},
booktitle={The Twelfth International Conference on Learning Representations},
year={2024},
url={https://openreview.net/forum?id=VtmBAGCN7o}
}

@misc{crewai2024,
  author = {Moura, João and Contributors},
  title = {CrewAI: Framework for orchestrating role-playing, autonomous AI agents},
  year = {2024},
  publisher = {GitHub},
  journal = {GitHub repository},
  howpublished = {\url{https://github.com/crewAIInc/crewAI}},
  commit = {Insert specific commit hash if applicable}
}

@misc{wang2024agentailanggraphmodular,
      title={Agent AI with LangGraph: A Modular Framework for Enhancing Machine Translation Using Large Language Models}, 
      author={Jialin Wang and Zhihua Duan},
      year={2024},
      eprint={2412.03801},
      archivePrefix={arXiv},
      primaryClass={cs.CL},
      url={https://arxiv.org/abs/2412.03801}, 
}

@inproceedings{cemri2025why,
  title={Why Do Multi-Agent {LLM} Systems Fail?},
  author={Cemri, Mert and Pan, Melissa Z. and Yang, Shuyi and Agrawal, Lakshya A. and Chopra, Bhavya and Tiwari, Rishabh and Keutzer, Kurt and Parameswaran, Aditya and Klein, Dan and Ramchandran, Kannan and Zaharia, Matei and Gonzalez, Joseph E. and Stoica, Ion},
  booktitle={Advances in Neural Information Processing Systems},
  volume={38},
  year={2025},
  url={https://proceedings.neurips.cc/paper_files/paper/2025/hash/b1041e52d3be19f0a9bc491657488e4a-Abstract-Datasets_and_Benchmarks_Track.html}
}

@misc{bai2026aiagentsspendmoney,
      title={How Do AI Agents Spend Your Money? Analyzing and Predicting Token Consumption in Agentic Coding Tasks}, 
      author={Longju Bai and Zhemin Huang and Xingyao Wang and Jiao Sun and Rada Mihalcea and Erik Brynjolfsson and Alex Pentland and Jiaxin Pei},
      year={2026},
      eprint={2604.22750},
      archivePrefix={arXiv},
      primaryClass={cs.CL},
      url={https://arxiv.org/abs/2604.22750}, 
}

@inproceedings{lin-etal-2026-agentask,
    title = "{A}gent{A}sk: Multi-Agent Systems Need to Ask",
    author = "Lin, Bohan  and
      Yang, Kuo  and
      Tan, Zelin  and
      Lai, Yingchuan  and
      Zhang, Chen  and
      Zhang, Guibin  and
      Yu, Xinlei  and
      Yu, Miao  and
      Wang, Xu  and
      Zhang, Yudong  and
      Wang, Yang",
    booktitle = "Proceedings of the 64th Annual Meeting of the {A}ssociation for {C}omputational {L}inguistics (Volume 1: Long Papers)",
    month = jul,
    year = "2026",
    address = "San Diego, California, United States",
    publisher = "Association for Computational Linguistics",
    url = "https://aclanthology.org/2026.acl-long.1294/",
    doi = "10.18653/v1/2026.acl-long.1294",
    pages = "28055--28077",
    ISBN = "979-8-89176-390-6"
}

@misc{jia2026masfirefaultinjectionreliability,
      title={MAS-FIRE: Fault Injection and Reliability Evaluation for LLM-Based Multi-Agent Systems}, 
      author={Jin Jia and Zhiling Deng and Zhuangbin Chen and Yingqi Wang and Zibin Zheng},
      year={2026},
      eprint={2602.19843},
      archivePrefix={arXiv},
      primaryClass={cs.SE},
      url={https://arxiv.org/abs/2602.19843}, 
}

@misc{wang2023augmentinglanguagemodelslongterm,
      title={Augmenting Language Models with Long-Term Memory}, 
      author={Weizhi Wang and Li Dong and Hao Cheng and Xiaodong Liu and Xifeng Yan and Jianfeng Gao and Furu Wei},
      year={2023},
      eprint={2306.07174},
      archivePrefix={arXiv},
      primaryClass={cs.CL},
      url={https://arxiv.org/abs/2306.07174}, 
}

@misc{zhang2024llmcascade,
  title={{LLM} Cascade with Multi-Objective Optimal Consideration},
  author={Zhang, Kai and Peng, Liqian and Wang, Congchao and Go, Alec and Liu, Xiaozhong},
  year={2024},
  eprint={2410.08014},
  archivePrefix={arXiv},
  primaryClass={cs.CL},
  url={https://arxiv.org/abs/2410.08014}
}

@misc{pei2025flowofaction,
  title={Flow-of-Action: {SOP} Enhanced {LLM}-Based Multi-Agent System for Root Cause Analysis},
  author={Pei, Changhua and Wang, Zexin and Liu, Fengrui and Li, Zeyan and Liu, Yang and He, Xiao and Kang, Rong and Zhang, Tieying and Chen, Jianjun and Li, Jianhui and Xie, Gaogang and Pei, Dan},
  year={2025},
  eprint={2502.08224},
  archivePrefix={arXiv},
  primaryClass={cs.AI},
  url={https://arxiv.org/abs/2502.08224}
}

@misc{alsayyad2026agenttraces,
      title={AgentTrace: A Structured Logging Framework for Agent System Observability}, 
      author={Adam AlSayyad and Kelvin Yuxiang Huang and Richik Pal},
      year={2026},
      eprint={2602.10133},
      archivePrefix={arXiv},
      primaryClass={cs.SE},
      url={https://arxiv.org/abs/2602.10133}, 
}

@inproceedings{itobservability,
author = {Silva Fontes, Gabriel and Andrikopoulos, Vasilios and Yumi Nakagawa, Elisa},
title = {Open Source Software Ecosystem for Cloud Observability: An Overview and Trends},
year = {2026},
isbn = {9798400723957},
publisher = {Association for Computing Machinery},
address = {New York, NY, USA},
url = {https://doi.org/10.1145/3786163.3788453},
doi = {10.1145/3786163.3788453},
booktitle = {Proceedings of the 14th IEEE/ACM International Workshop on Software Engineering for Systems-of-Systems and Software Ecosystems},
pages = {9–15},
numpages = {7},
location = {
},
series = {SESoS '26}
}

@misc{solomon2025lumimas,
      title={LumiMAS: A Comprehensive Framework for Real-Time Monitoring and Enhanced Observability in Multi-Agent Systems},
      author={Ron Solomon and Yarin Yerushalmi Levi and Lior Vaknin and Eran Aizikovich and Amit Baras and Etai Ohana and Amit Giloni and Shamik Bose and Chiara Picardi and Yuval Elovici and Asaf Shabtai},
      year={2025},
      eprint={2508.12412},
      archivePrefix={arXiv},
      primaryClass={cs.CR},
      url={https://arxiv.org/abs/2508.12412},
}

@misc{he2025sentinelagent,
      title={SentinelAgent: Graph-based Anomaly Detection in Multi-Agent Systems},
      author={Xu He and Di Wu and Yan Zhai and Kun Sun},
      year={2025},
      eprint={2505.24201},
      archivePrefix={arXiv},
      primaryClass={cs.AI},
      url={https://arxiv.org/abs/2505.24201},
}

@misc{moshkovich2025taming,
      title={Taming Uncertainty via Automation: Observing, Analyzing, and Optimizing Agentic AI Systems},
      author={Dany Moshkovich and Sergey Zeltyn},
      year={2025},
      eprint={2507.11277},
      archivePrefix={arXiv},
      primaryClass={cs.AI},
      url={https://arxiv.org/abs/2507.11277},
}

\appendix
\section{Appendix}

\subsection{Metric Definitions and Computation} \label{app:metrics}

 A query/context is identified by its \textit{context\_id}. Metrics are computed over recent observation windows, with default windows of 20 queries, 50 calls per tool, and 50 calls per worker agent unless otherwise specified.

\paragraph{Agent loop index.}
For execution context $c$, with $\mathcal{A}$ denoting the set of worker agents, the agent loop index is

\begin{equation}
L(c)=
\frac{N_{\mathrm{repeated}}(c)}
{N_{\mathrm{calls}}(c)},
\end{equation}

where $N_{\mathrm{calls}}(c)$ is the number of worker-agent calls and $N_{\mathrm{repeated}}(c)$ is the number of calls to an agent after its first occurrence within $c$. Equivalently,

\begin{equation}
N_{\mathrm{repeated}}(c) =
\sum_{a\in\mathcal{A}}
\max(N_a(c)-1,0).
\end{equation}

The metric is zero when no worker-agent calls are observed. A high value indicates repeated activation but does not establish that the repetition is erroneous.

\paragraph{Average steps.}
For query $q$, execution steps are defined as

\begin{equation}
S(q)=N_{\mathrm{agent}}(q)+N_{\mathrm{tool}}(q),
\end{equation}

where the two terms count worker-agent calls and tool invocations. The window-level average is

\begin{equation}
\overline{S}=
\frac{1}{|Q|}
\sum_{q\in Q}S(q).
\end{equation}

ResonAct uses this metric to identify executions whose step count exceeds the expected workload distribution. The current implementation uses the empirical $P_{95}$ as the upper reference for unbounded execution.

\subsection{Execution and Path Efficiency}

\paragraph{Unnecessary path ratio.}
The unnecessary path ratio provides a proxy for wasted execution by combining repeated worker-agent calls and failed tool invocations:

\begin{equation}
U(q)=
\frac{
N_{\mathrm{repeated\ agent}}(q)+
N_{\mathrm{failed\ tool}}(q)}
{
N_{\mathrm{agent}}(q)+N_{\mathrm{tool}}(q)
}.
\end{equation}

The metric is zero when no counted execution occurs. It is intentionally a heuristic measure: legitimate repeated agent calls may be classified as unnecessary, and repeated successful tool calls do not contribute to the numerator.

\paragraph{Delegation efficiency.}
Delegation efficiency measures successful worker completion relative to delegation activity:

\begin{equation}
D(q)=
\frac{N_{\mathrm{successful\ worker}}(q)}
{N_{\mathrm{delegated}}(q)}.
\end{equation}

Here, $N_{\mathrm{delegated}}$ counts worker-agent delegation events and $N_{\mathrm{successful\ worker}}$ counts worker completions with successful outcomes. The metric provides evidence about the effectiveness of information and work transfer across agent boundaries, but does not match individual assignments to completions or independently verify answer quality.

\paragraph{Compactness.}
Compactness measures execution length relative to the shortest positive-step completed context in the current window. Let

\begin{equation}
S_{\min}=
\min_{q\in Q_{\mathrm{completed}},S(q)>0}S(q).
\end{equation}

For a completed context $q$ with $S(q) > 0$,

\begin{equation}
C(q)=
\frac{S_{\min}}{S(q)}.
\end{equation}

The value is zero when the context is incomplete, has no steps, or no positive-step completed baseline exists. Compactness therefore measures relative execution efficiency rather than optimality.

\paragraph{API call efficiency.}
API call efficiency measures recent tool-call volume relative
to a longer moving baseline. Let $Q_s \subseteq Q_\ell$ denote
the short and long windows of execution contexts, ordered by
their most recent update time. Using $N_{\mathrm{tool}}(q)$
for the number of tool invocations in context $q$, define
\begin{equation}
\mu_s = \frac{1}{|Q_s|}
        \sum_{q \in Q_s} N_{\mathrm{tool}}(q),
\qquad
\mu_\ell = \frac{1}{|Q_\ell|}
           \sum_{q \in Q_\ell} N_{\mathrm{tool}}(q).
\end{equation}
For $\mu_\ell > 0$, the metric is
\begin{equation}
E_{\mathrm{API}} = \frac{\mu_s}{\mu_\ell}.
\end{equation}
The short and long windows contain up to 5 and 20 contexts,
respectively, by default; available contexts are used while
the history accumulates. When $\mu_\ell = 0$, the metric is
reported as unavailable.

Values above 1 indicate increased recent tool-call volume
relative to the longer baseline, while values below 1
indicate reduced volume. The metric is a drift signal:
it does not independently establish task success, monetary
cost, or execution optimality. Changes in task difficulty
can also change the ratio.


\subsection{Agent and Tool Health}

\paragraph{Tool health and latency.}
For tool $t$, the failure rate over its observation window is

\begin{equation}
F_t=
\frac{N_{\mathrm{failed}}(t)}
{N_{\mathrm{terminal}}(t)},
\end{equation}

with success rate $1-F_t$. The latest status is \textit{up} for a successful completion and \textit{down} for a failed completion. Tool latency is measured from invocation to terminal completion or failure:

\begin{equation}
\ell_t=
T_{\mathrm{terminal}}-T_{\mathrm{invocation}}.
\end{equation}

ResonAct considers health and latency jointly because a low latency can also result from a rapid failure.

\paragraph{Agent health and latency.}
For worker agent $a$, health is defined from recent completion outcomes:

\begin{equation}
F_a=
\frac{N_{\mathrm{failed\ completions}}(a)}
{N_{\mathrm{completions}}(a)}.
\end{equation}

Agent latency is obtained from the recorded duration of completed worker-agent executions. Health and latency provide complementary signals for detecting sustained agent degradation. The host/orchestrator is excluded from these metrics.

\subsection{Semantic and Quality Signals}

\paragraph{Tool-routing confidence.}
For a selected tool $t^*$ and query $q$, ResonAct computes semantic similarity between the query and each tool schema using cosine similarity:

\begin{equation}
s_t=
\cos
\left(
E(q),E(\mathrm{schema}_t)
\right).
\end{equation}

The selected-tool confidence is then obtained through a temperature-scaled softmax:

\begin{equation}
P(t^*\mid q)=
\frac{\exp(s_{t^*}/T)}
{\sum_{t\in\mathcal{T}}\exp(s_t/T)}.
\end{equation}

The current implementation uses the \texttt{all-MiniLM-L6-v2} embedding model with a default temperature of $T=0.1$. The score reflects how strongly the selected tool semantically matches the query relative to the other tools in the registry; it is not a calibrated probability of correctness.

\paragraph{LLM-judge scores.}
For completed queries, ResonAct optionally uses an LLM judge to score communication quality, planning quality, coordination, and final-answer quality. Communication, planning, and answer quality are independently scored on a 1--5 scale. Coordination is computed as

\begin{equation}
J_{\mathrm{coord}}=
\frac{
J_{\mathrm{communication}}+
J_{\mathrm{planning}}
}{2}.
\end{equation}

Scores may be normalized by dividing by 5. These metrics provide qualitative evidence about execution quality but are model-based assessments rather than independently verified correctness labels.

\paragraph{Agent activation accuracy.}
Agent activation accuracy evaluates whether the observed ordered agent/tool execution path belongs to a set of accepted paths associated with the query intent:

\begin{equation}
A =
\frac{N_{\text{utterances with accepted paths}}}
     {N_{\text{scored utterances}}}.
\end{equation}

The implementation groups queries by semantic intent using MiniLM embeddings and a default similarity threshold of 0.75. During bootstrapping, an LLM may provide an optimal reference path. Once sufficient observations are collected, frequently observed paths are incorporated into the accepted-path set. The default bootstrap threshold is 50 utterances per intent, and the retained path probability mass is 0.80. Consequently, the metric measures agreement with a learned execution-path reference rather than accuracy against an independently labeled ground truth.

These metrics are computed continuously and aggregated at the context, agent, and system levels.

\end{document}